\documentclass[runningheads]{llncs}
\usepackage[T1]{fontenc}
\usepackage{graphicx}
\usepackage{multirow}
\begin{document}
\title{Fast calibration for ultrasound imaging \\guidance based on depth camera}
%
%
\author{Fuqiang Zhao\inst{1} \and
Mingchang Li\inst{2} \and
Mengde Li\inst{3} \and
Zhongtao Fu\inst{4} \and  
Miao Li\inst{1,3}
}
\authorrunning{F. Zhao et al.}
%
\institute{The School of Power and Mechanical Engineering, Wuhan University, China \and
Department of Neurosurgery, Renmin Hospital of Wuhan University, China\\\and
The Institute of Technological Sciences, Wuhan University, China\\ \and
School of Mechanical and Electrical Engineering, Wuhan Institute of Technology, Wuhan, China \\
\email{miao.li@whu.edu.cn}
}
%
\maketitle              
\begin{abstract}
During the process of robot-assisted ultrasound(US) puncture, it is important to estimate the location of the puncture from the 2D US images. To this end, the calibration of the US image becomes an important issue. In this paper, we proposed a depth camera-based US calibration method, where an easy-to-deploy device is designed for the calibration. With this device, the coordinates of the puncture needle tip are collected respectively in US image and in the depth camera, upon which a correspondence matrix is built for calibration. Finally, a number of experiments are conducted to validate the effectiveness of our calibration method. 
\footnote{\noindent This work was supported by Suzhou Key Industry Technology Innovation Project under the grant agreement number SYG202121. (Corresponding author:  Miao Li.)}

\keywords{Robot ultrasound  \and Robot calibration \and Robot-assisted puncture.}
\end{abstract}
\section{Introduction}
\begin{figure}[!h]
    \centering
    \includegraphics[width=0.8\linewidth]{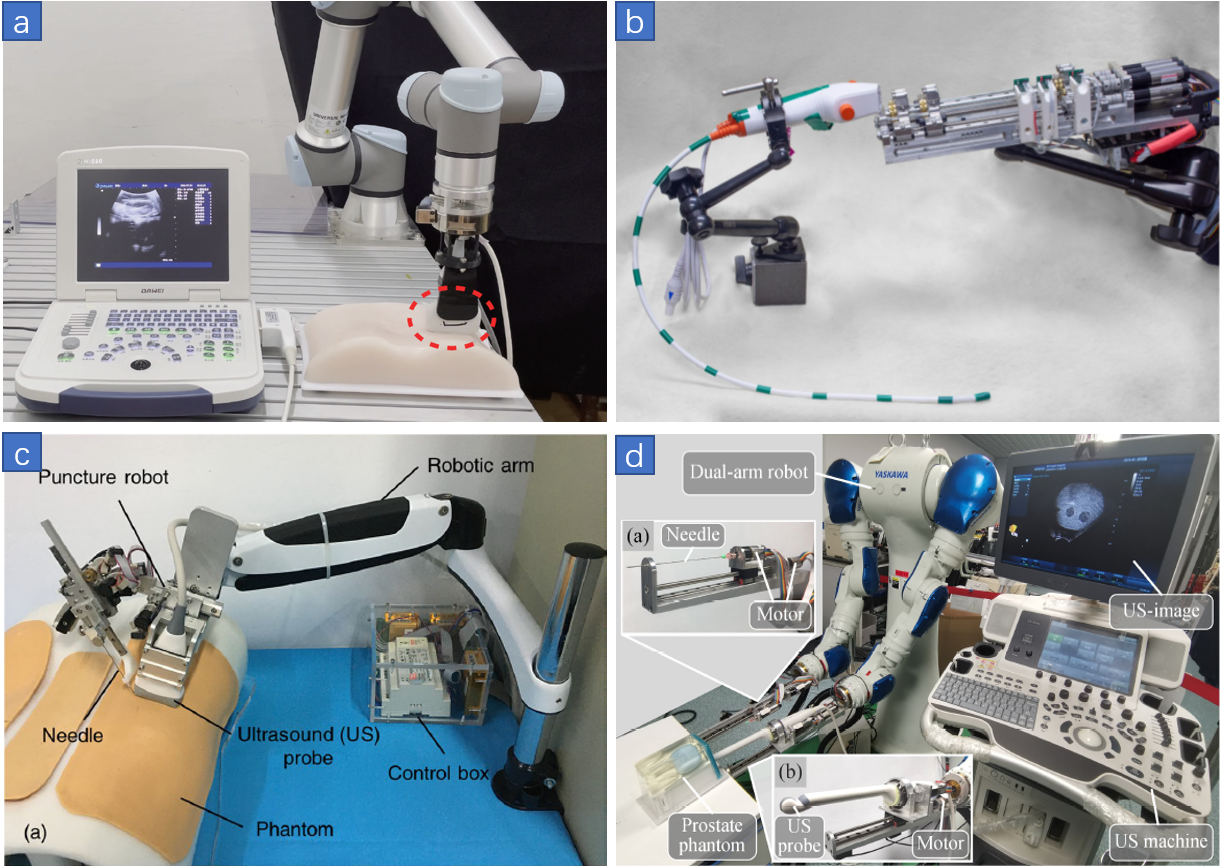}
    \caption{US-guided robotic puncture classification:(a) Independent US robot with manual biopsy \cite{deng2021learning}. (b) Independent puncture robot with manual US scanning \cite{hoelscher2021backward}. (c) Single-arm mechanical ultrasound-guided robotic puncture \cite{chen2021ultrasound}. (d) Dual-arm ultrasound-guided robotic puncture \cite{xiong2021mechanism}}.
    \label{classification}
    \vspace{-1cm}
\end{figure}

\begin{figure}[h]
    \centering
    \includegraphics[width=0.7\linewidth]{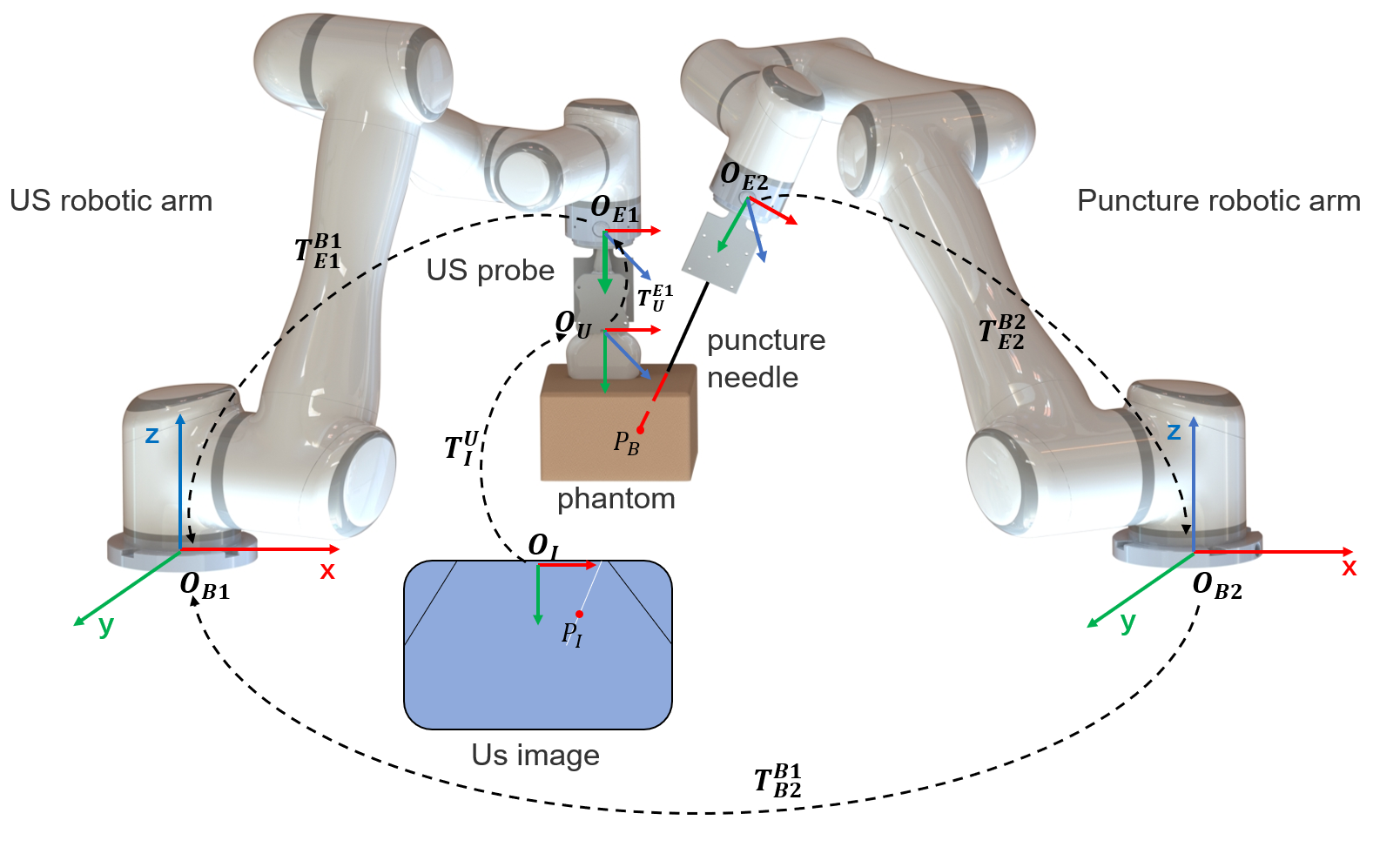}
    \caption{The dual-arm ultrasound-guided robotic puncture system consists of two robotic arms carrying an ultrasound probe and a puncture needle, respectively.} 
    \label{robotic2}
    \vspace{-0.5cm}
\end{figure}

With the fast development of ultrasound (US) imaging and medical robots \cite{taylor2022surgical}, US-guided robotic puncture has been widely studied recently \cite{podder2010mirab,stoianovici2012endocavity,su2014piezoelectrically,chen2016robotic,sarli2019turbot,xiong2021mechanism}. Compared with free-hand operation, US-guided robotic puncture can offer high accuracy and stability. In addition, the robots can achieve flexible control within a small surgical space and reduce surgical trauma. In general, 
ultrasound-guided robotic puncture can be classified into the following four categories: (1) Independent US robot with manual biopsy by a doctor. (2) Independent puncture robot with manual US scanning by a doctor. (3) Single-arm mechanical ultrasound-guided puncture robot. (4) Dual-arm ultrasound-guided robot puncture as shown in Fig.\ref{classification}. For all of these four scenarios, it is necessary to estimate the location of the puncture needle from the pure 2D ultrasound imaging. Especially for the dual-arm robot system, with the US probe and puncture needle installed on two separated arms, it is more crucial to know the precise location of the puncture needle as shown in Fig.\ref{robotic2}. In this work, we propose a new calibration method together with the auxiliary device for ultrasound imaging guidance task.


\subsection{Related Work}
To meet the clinical requirements, many US image calibration methods have been developed, which can be roughly divided into two categories, mechanism-based calibration and image-based calibration \cite{aalamifar2016robot,huang2018fully,li2012image}. 
The mechanism-based calibration method is mainly used for robotic system calibration \cite{wang2014towards}, while The image-based calibration method is the mainly used for US probe calibration\cite{wang2018automated}. 
Traditional US calibration methods require a phantom to provide a set of fiducials, such as the N-wire phantom method \cite{carbajal2013improving}. Carbajal et al. proposed an improved N-wire phantom freehand US calibration method based on the middle wires to improve the calibration accuracy \cite{carbajal2013improving}. To reduce the isotropic fiducial localization errors in this method, Najafi et al. proposed a multiwedge phantom calibration method to achieve higher calibration accuracy \cite{najafi2014closed}. In this method, the calibration matrix was solved with a closed-form solution, which enables easy and accurate US calibration. Afterward, Shen et al. proposed a new method considering the use of wires for US calibration \cite{shen2019method}. However, these methods need external tracker and wire or wedge phantom, which also accumulate errors from phantom. Moreover, the process of mounting and dismounting the phantom will reduce efficiency, and increase the complexity of calibration process \cite{kim2013ultrasound}.

To address these issue, Hunger et al. attached a small rubber sleeve and a softball to the needle tip \cite{hungr20123}, which improved the positioning of the needle tip from the US images. However, the softball itself also resulted in extra errors. Recently, Xiong et al. proposed a calibration method based on mechanism-image fusion for an ultrasound-guided two-arm robot \cite{xiong2021mechanism}. The pixel position of the needle tip in the US image is calculated by manual annotation, which also reduces the efficiency of registration. Due to the presence of US noise and artifacts, calibration accuracy and efficiency is still challenging in this approach.

 In this paper, we proposed a fast calibration method based on depth camera.A puncture needle is used to replace the reference phantoms in previous works, which could reduce the extra errors from calibration model. A depth camera is used to annotation the needle tips automatically. The coordinate system transformation between the depth camera and the US images can be used to formulate the calibration process. Experimental results shows that our approach achieve better performance compared with previous methods \cite{carbajal2013improving} and \cite{pagoulatos2001fast}.
%
\section{Theory and Method}
\subsection{Calibration Setup}
\begin{figure}[!h]
    \centering
    \includegraphics[width=0.8\linewidth]{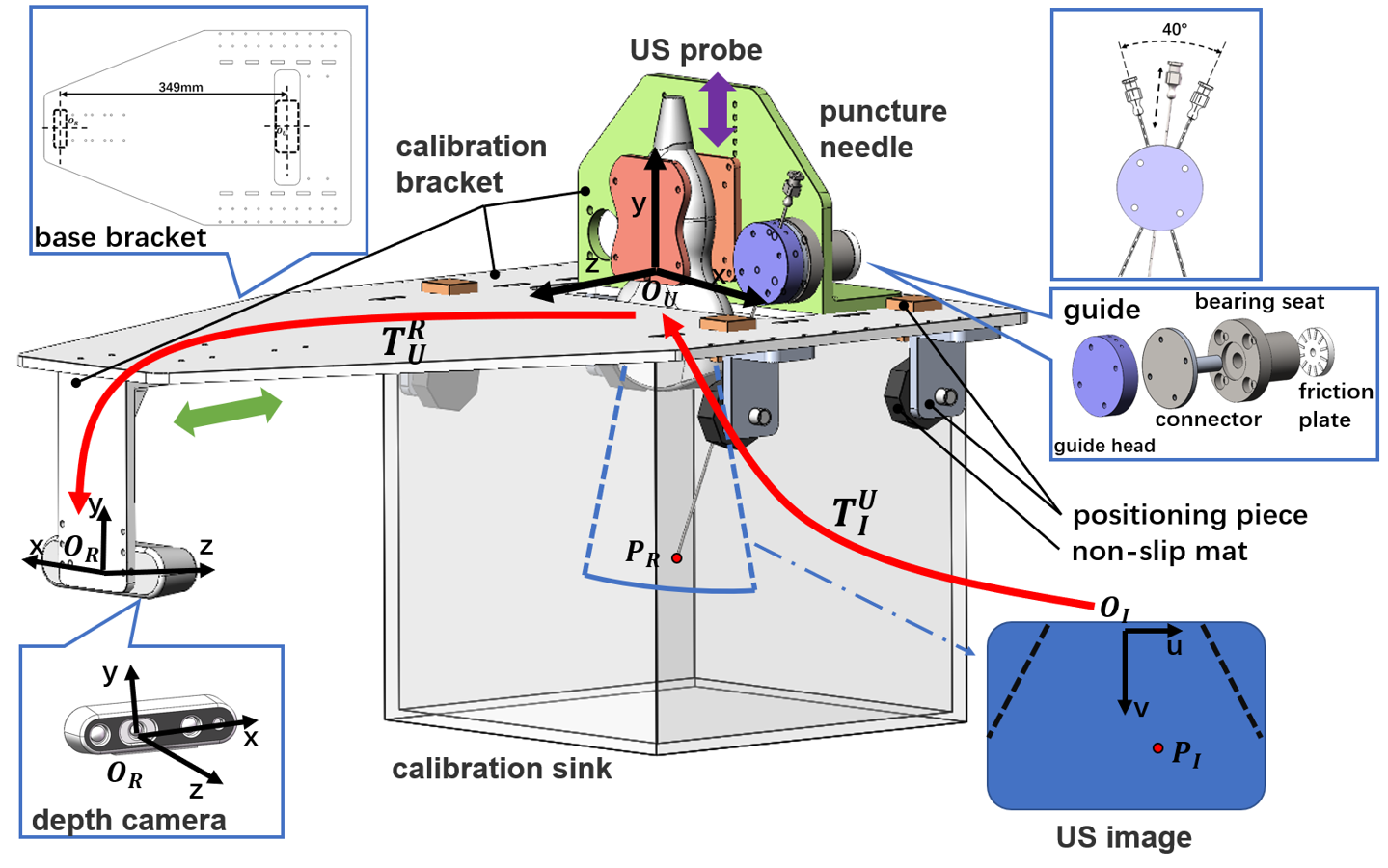}
    \caption{US Calibration Setup. It consists of an US system, a depth camera, a puncture needle, a calibration sink, and a calibration phantom bracket.}
    \label{Model}
    \vspace{-0.5cm}
\end{figure}

The US image calibration system is shown in Fig.~\ref{Model}, which consists of an US system, a depth camera, a puncture needle, a calibration sink, and a calibration phantom bracket. 
The medical US imaging uses the attenuation rate of sound waves to detect the internal structure of entities. Therefore, US imaging can be simplified as a linear model, and various parameters of US system can be obtained through calibration. The US image calibration system utilizes this linear model and coordinate system transformation, which means that every point in the US image has a corresponding point in the world coordinate system. 

Before the calibration, the calibration bracket is placed on the calibration sink. The position can be adjusted with the positioning piece, as shown in Fig.~\ref{Model}. Multiple sets of positioning holes for the positioning bracket are designed on the base, and these evenly arranged positioning holes can adjust the position of the positioning bracket.

The coordinate system of the depth camera is defined as the world coordinate system, and the optical center of the depth camera is set as the origin of the actual coordinate system.
 The puncture needle moves along the guide, which is connected to the bearing and can rotate to adjust the puncture needle angle. In Fig.~\ref{Model}, the green arrow indicates the movable direction of the depth camera positioning bracket and purple arrow indicates the movable direction of the US probe positioning bracket. Note that if the probe is mounted on the robotic arm, these movement can be accomplished by the robotic arm.

\subsection{Calibration Process}
\begin{figure}[!h]
    \centering
    \includegraphics[width=0.7\linewidth]{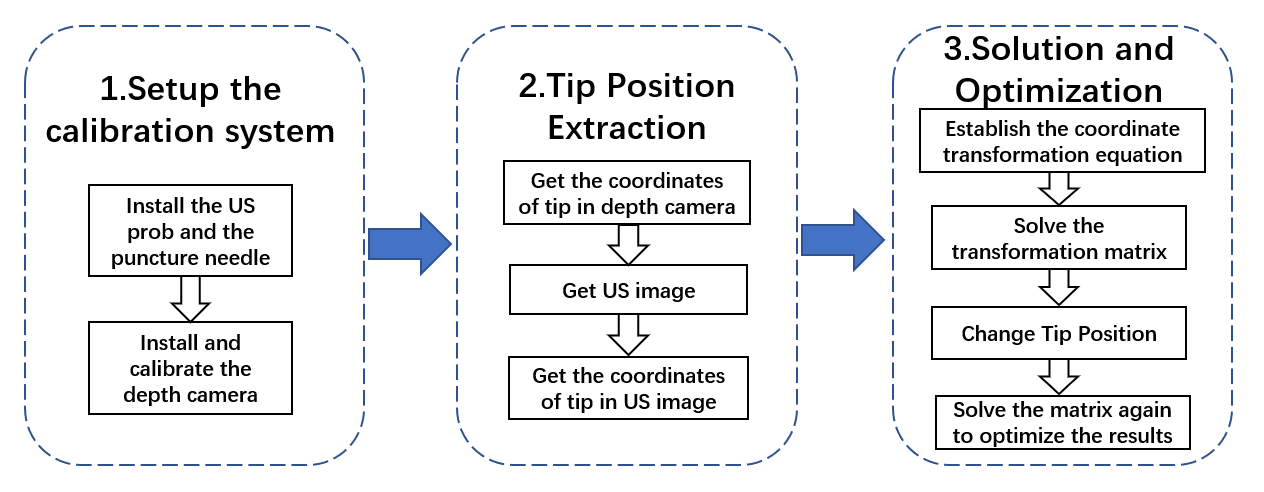}
    \caption{US calibration process includes: system setup, date collection and optimization.}
    \label{process}
    \vspace{-0.5cm}
\end{figure}

Before calibration, it is necessary to establish the coordinate systems for the depth camera and US probe, and determine the mapping relationship between the coordinate systems based on the structural parameters of the calibration bracket. In the overall framework of US image calibration, three coordinate systems are defined: depth camera coordinate system, US probe coordinate system, and US image coordinate system, represented by R, U, and I, respectively. The two corresponding coordinate transformation matrices are $T_ U ^ R$ and $T_ I ^ U$, which actually represent the external and internal parameters in the US image calibration process.

The registration process shown in Fig.\ref{process} consists of the following steps: (1) Install the depth camera and US probe on the base through their respective positioning brackets, and install the puncture needle on the calibration bracket through guide components and bearings to build a calibration system. (2) Establish the coordinate systems, set the coordinate system of the depth camera to $O_R$, and the coordinate origin is the optical center of the depth camera. The XOY plane is parallel to the imaging plane of the depth camera. The coordinate system of the US probe is $O_U$, and the origin of the coordinate is the center of the US probe. The XOY plane is parallel to the US imaging plane. The coordinate system of the US image is set to $O_I$, and the coordinate origin is the top-point of the center-line of the US image. The UOV plane coincides with the US image plane, as shown in Fig.~\ref{Model}.

\subsection{Segmentation and Localization of Needle Tip}
\begin{figure}[h]
    \centering
    \includegraphics[width=0.6\linewidth]{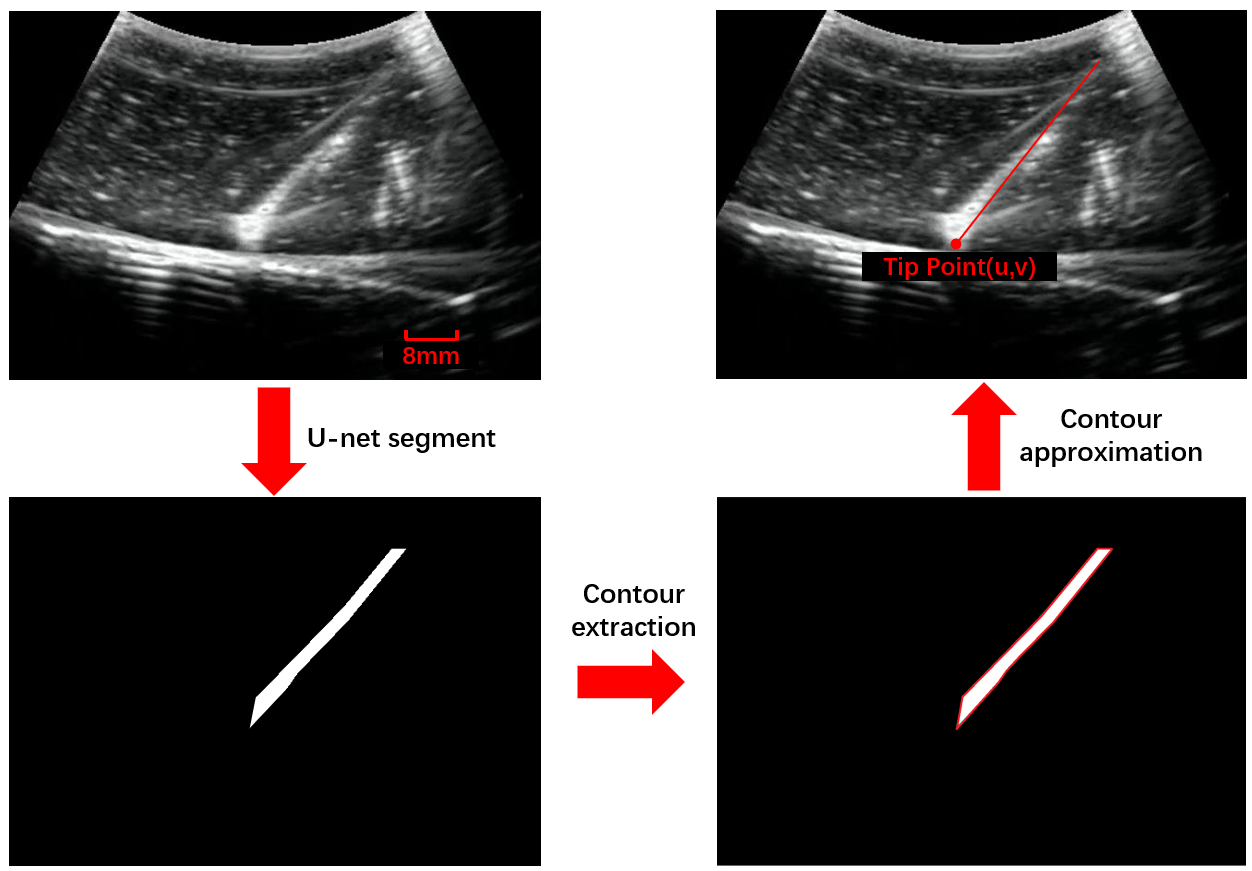}
    \caption{Needle tip positioning process} 
    \label{tip}
    \vspace{-0.5cm}
\end{figure}

In order to accurately locate the needle tip position in US images, we use U-Net neural network to train the US images and to recognize the needle. U-Net neural network is a widely used image segmentation network framework for feature extraction from medical images \cite{ronneberger2015u}. 
In this work, we collect $200$ US images of a puncture needle in water using an US system.The outline of the puncture needle in the images are annotated as the training set. After training the U-net network, we can extract the needle tip contour segmentation in the US images. Then, the needle is fitted into line segments and the end of the line segment that enters the image first according to the needle insertion direction is selected as the needle tip coordinate. The process of the needle tip segmentation and localization process is shown in Fig.~\ref{tip}.

\subsection{Calibration Model}
The spatial mapping of a point from an US image to a depth camera is represented as:
\begin{equation}
P_R=T \cdot P_I=T_U^R \cdot T_I^U \cdot P_I
\end{equation}
$P_I$ represents the position of a point in the US image, $P_I=[u,~v,~1]^\mathrm{T}$. $u$ and $v$ are the  coordinates of the point in the US image with the unit of pixel. $P_R$ represents the position of the point in the depth camera, $P_R=[x_r, ~y_r, ~z_r, ~1]^\mathrm{T}$, $x_r$, $y_r$, $z_r$ are the coordinates of the point in the depth camera. In order to ensure the accuracy of calibration and reduce the requirements for calibration system, we use the needle tip coordinates of the puncture needle to calculate the transformation matrix $T$.

Transformation matrix $T_U^R$ represents the transformation matrix between the depth camera and the US probe, which is also the extrinsic matrix of the US probe and determines the position of the US image imaging plane. Based on the structure of the designed calibration bracket, we can easily obtain the transformation matrix parameters $x_u$, $y_u$, $z_u$ between the depth camera positioning bracket and the US probe positioning bracket. The transformation matrix is as follows:

\begin{equation}
T_U^R
=
\left[
\begin{array}{cccc}
1 &~~0 &~~0 &~~x_u \\
0 &~~1 &~~0 &~~y_u \\
0 &~~0 &~~1 &~~z_u \\
0 &~~0 &~~0 &~~1
\end{array}
\right]
\end{equation}

Transformation matrix $T_I^U = [I_x, I_y, I_0]$ represents the transformation matrix between the US probe and the US image, which is the intrinsic matrix of the US probe. The three vectors $I_0$, $I_x$ and $I_y$ here each contain four scalars to represent the main calibration parameters. 




It should be noted that in order to solve $T_I^U$, we need to use the puncture needle guide on the calibration bracket to change the position of the needle tip multiple times to obtain a sufficient number of reference points. For ease of calculation, we select any three non-collinear points $P_{R1}$, $P_{R2}$, $P_{R3}$, and combine their coordinates under the depth camera into a matrix $P_{R'}$, represented as:

\begin{equation}
P_{R'}=
\left[
\begin{array}{ccc}
P_{R1} &~P_{R2} &~P_{R3}
\end{array}
\right]
=
\left[
\begin{array}{ccc}
x_{r1} &~~x_{r2} &~~x_{r3}\\
y_{r1} &~~y_{r2} &~~y_{r3}\\
z_{r1} &~~z_{r2} &~~z_{r3}\\
1 &~~1 &~~1 
\end{array}
\right]
\end{equation}

Similarly, the coordinates in US images are combined as matrix $P_{I'}$, represented as:
\begin{equation}
P_{I'}=
\left[
\begin{array}{ccc}
P_{I1} &~P_{I2} &~P_{I3}
\end{array}
\right]
=
\left[
\begin{array}{ccc}
u_{i1} &~~u_{i2} &~~u_{i3}\\
v_{i1} &~~v_{i2} &~~v_{i3}\\
1 &~~1 &~~1 
\end{array}
\right]
\end{equation}
The spatial mapping of US image to depth camera becomes:
\begin{equation}
P_{R'}=T \cdot P_{I'}=T_U^R \cdot T_I^U \cdot P_{I'}
\end{equation}
Based on (1)(5)(6)(7), matrix operation $T_I^U$ can be represented as follows and the $inv$ represents the generalized inverse of the matrix.

\begin{equation}
T_I^U=inv(T_U^R) \cdot P_{R'} \cdot inv(P_{I'})
\label{eqn:cal}
\end{equation}

\section{Experiments and Results}
\subsection{Experiment Setup}
\begin{figure}[h]
    \centering
    \includegraphics[width=0.6\linewidth]{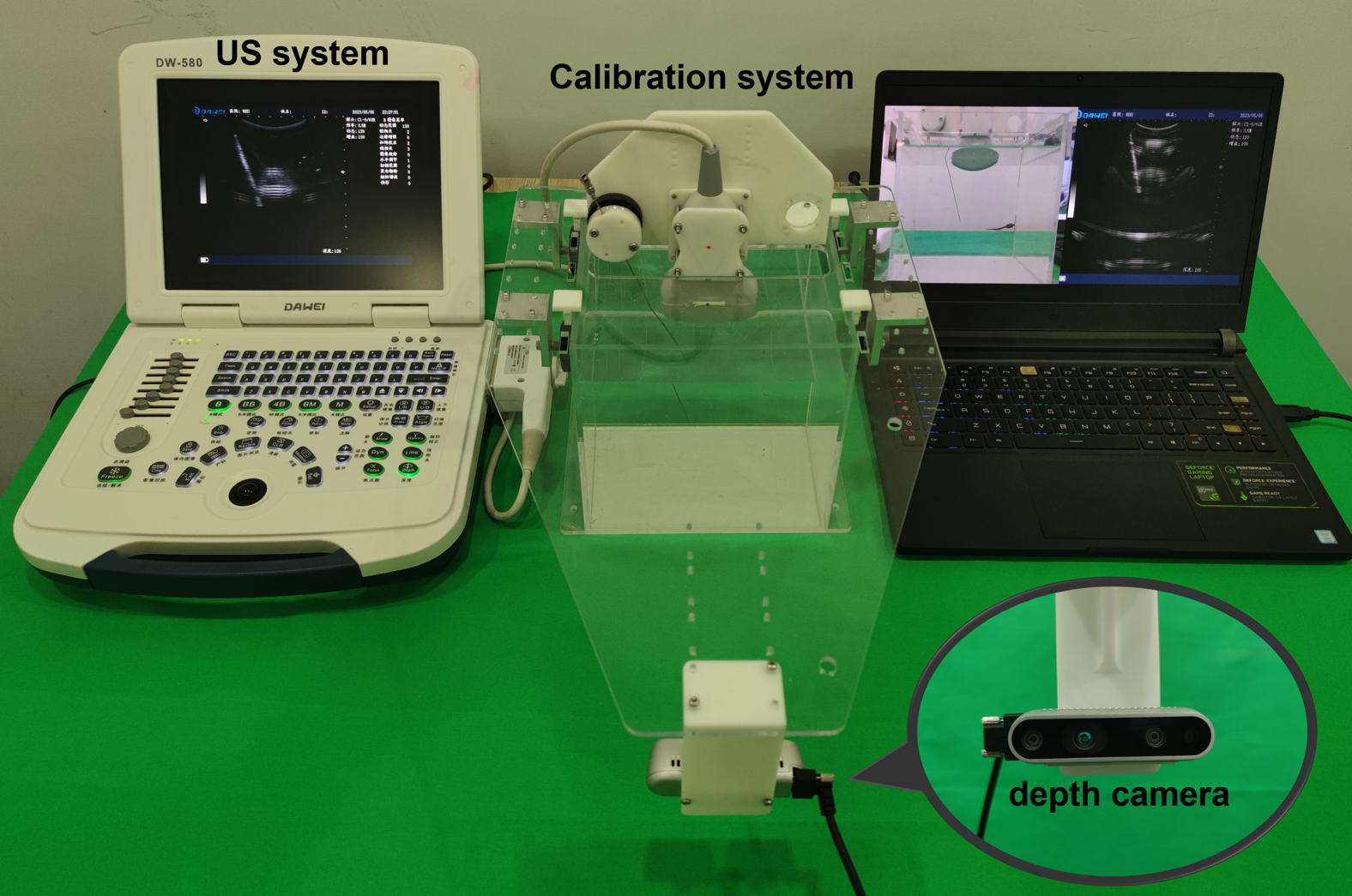}
    \caption{US calibration experimental equipment} \label{equipment}
    \vspace{-0.5cm}
\end{figure}

In order to verify the accuracy of the US image calibration system, experiments were conducted on our proposed US calibration system as shown in Fig.~\ref{equipment}. The base bracket in the calibration bracket is made of acrylic plate, and the positioning brackets for the depth camera and US probe, as well as the guide component for the puncture needle, are made of high-precision 3D printing and processing. We select Realsense-D435i for the depth camera, DW-580 for the US system, and the puncture needle diameter is of $0.8$ mm.

\subsection{Position Extraction of Coordinate System and Needle Tip}
We establish the coordinate system for the depth camera, US probe, and US image. The transformation matrix $T_U^R$ between the depth camera and the US probe is given based on the structural parameters of the calibration bracket:
\begin{center}
${T_U^R
=
\left[
\begin{array}{cccc}
1 &~~0 &~~0 &~~35.31 \\
0 &~~1 &~~0 &-50.24 \\
0 &~~0 &~~1 &~~349.00 \\
0 &~~0 &~~0 &~~1
\end{array}
\right]
}$
\end{center}

\begin{figure}[h]
\centering
\includegraphics[width=0.6\linewidth]{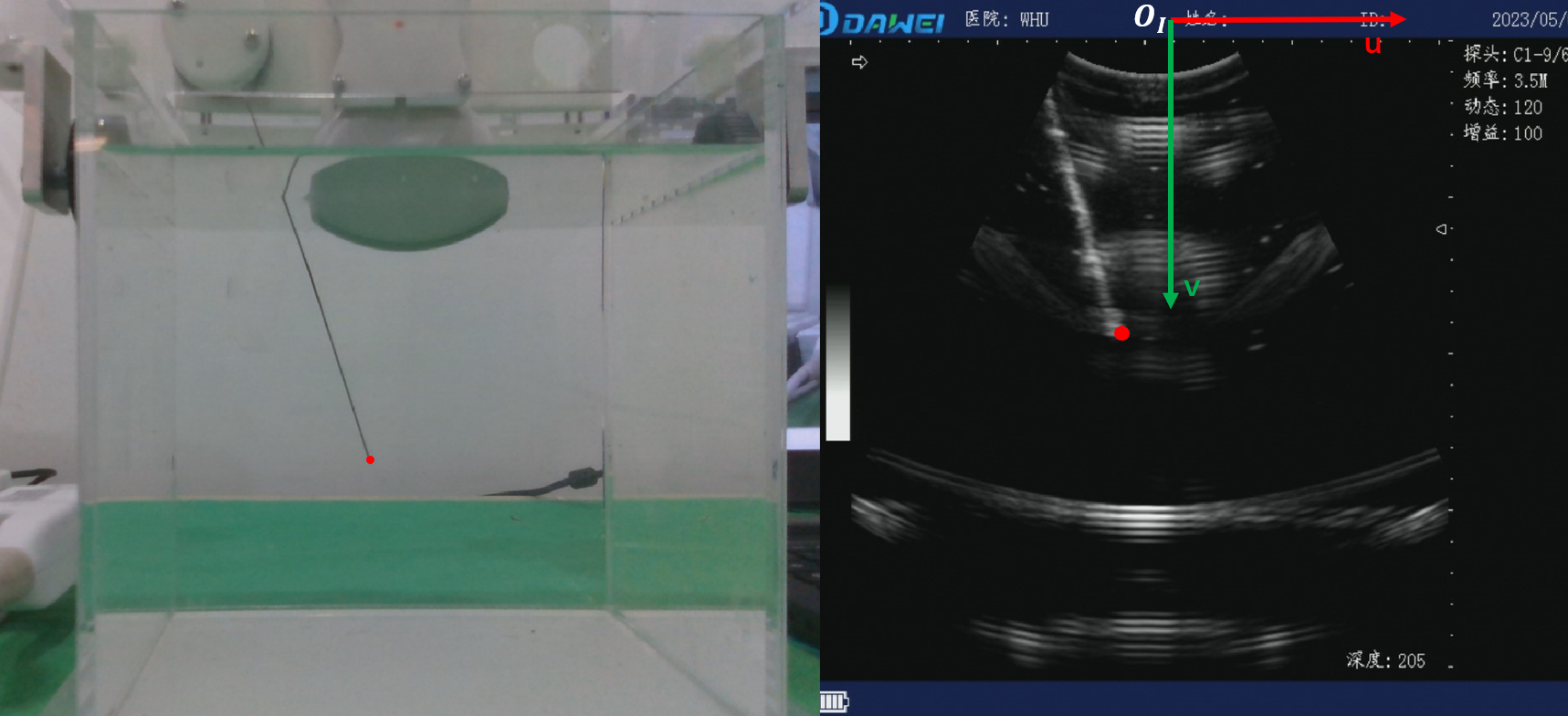}
\caption{Needle tip in depth camera image (left) and in the US image (right)}
\label{tipimage}
\end{figure}

For each set of needle coordinates, it is necessary to obtain the coordinates of the needle in the depth camera and the US image. According to the equation, at least three pairs of needle coordinates are needed to calculate the transformation matrix. To ensure accuracy and reduce errors, it is possible to collect more pairs of needle coordinates.

When obtaining the coordinates of the needle tip in the depth camera, painting fluorescent markers on the needle tip can increase localization accuracy, and also reduce the positioning error caused by the reflection of the puncture needle due to poor lighting condition. 

Due to the severe attenuation of US in the air, it is necessary to fill the calibration sink with water when using the US system, the US probe can clearly obtain the image of the needle tip in the sink. By using the U-net network to segment the US image of the needle tip, the coordinates of the needle tip in the US image can be accurately located. The position of the needle tip in the US and depth camera is shown in Fig.~\ref{tipimage}.

\begin{figure}[h]
    \centering
    \includegraphics[width=0.8\linewidth]{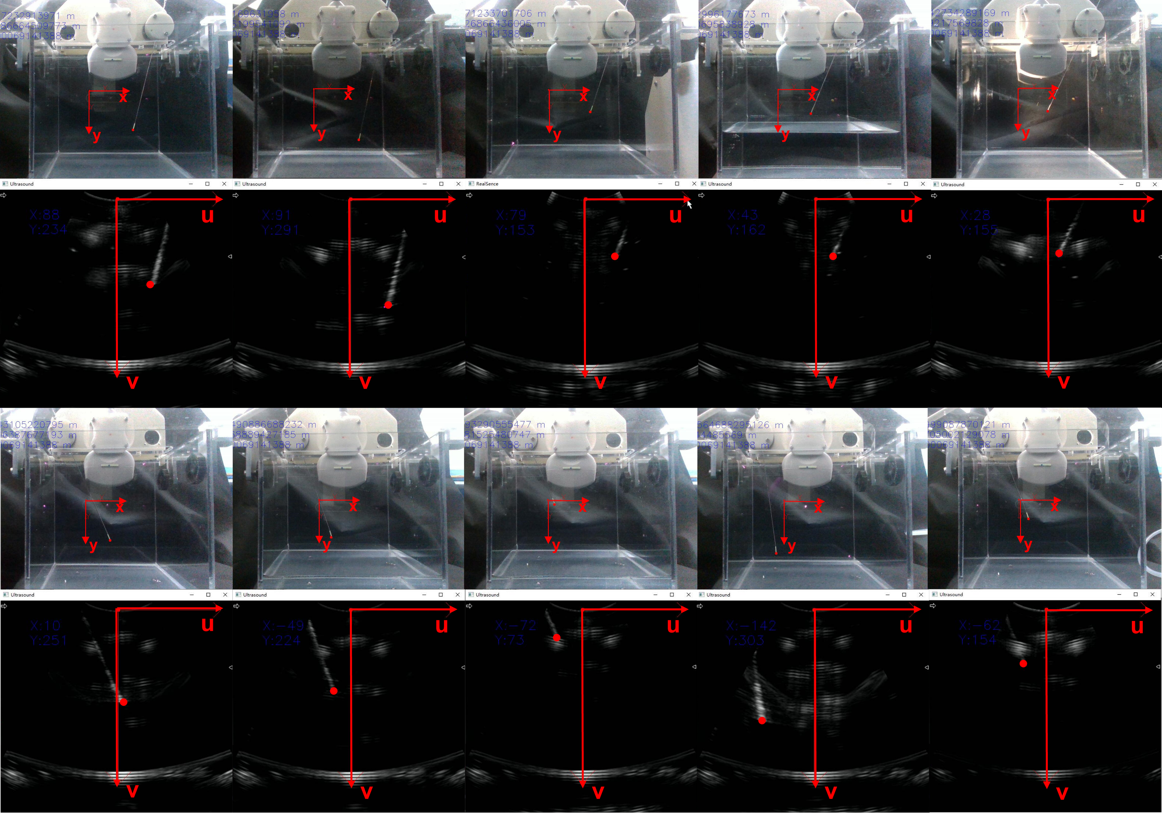}
    \caption{Experimental collection of images} \label{date}
\end{figure}

\textbf{Remarks:} It should be noted that when using a depth camera to obtain the needle coordinates, the water in the sink needs to be extracted, as there will be obvious refraction phenomena in the water. This can lead to significant positioning errors. We sampled $10$ pairs of points, and their coordinate information in depth camera and US images, as shown in Fig.~\ref{date}.

\subsection{Calibration Results}
Using the collected data points, the final solution of $T_I^U$ is obtained using equation (\ref{eqn:cal}). After obtaining the US intrinsic matrix, the only corresponding point in the depth camera can be calculated through coordinate transformation for any point in any US image.

\begin{center}
${T_I^U
=
\left[
\begin{array}{ccc}
0.3418 &~~0.0074 &~-0.6193 \\
-0.0025 &~~0.3502 &~~28.2740 \\
0 &~~0 &~~0 \\
0 &~~0 &~~1
\end{array}
\right]
}$
\end{center}

\begin{table}[h]
\caption{Results of US image calibration.}\label{tab1}
\tabcolsep=0.62em
\renewcommand\arraystretch{1.2}
\begin{tabular}{cccccccccc}
\hline
\multirow{2}{*}{\textbf{No}} & \multicolumn{2}{c}{$P_I$} & \multicolumn{3}{c}{Physical $P_W(mm)$} & \multicolumn{3}{c}{Calibration $P_R(mm)$} & Errors(mm)      \\ \cline{2-9}
                                 & u          & v         & x          & y         & z          & x            & y           & z         & $|P_R-P_W|$ \\ \hline
1                                & 88         & 234       & 66.40      & 59.94     & 349.0     & 66.488       & 59.764      & 349.0     & 0.1968      \\
2                                & 91         & 291       & 68.70      & 77.60     & 349.0    & 67.933       & 79.719      & 349.0     & 2.2535      \\
3                                & 79         & 153       & 62.18      & 33.07     & 349.0     & 62.817       & 31.420      & 349.0     & 1.7686      \\
4                                & 43         & 162       & 49.51      & 35.52     & 349.0     & 51.604       & 34.654      & 349.0     & 2.2660      \\
5                                & 28         & 155       & 44.14      & 33.45     & 349.0     & 45.400       & 32.247      & 349.0     & 1.7421      \\
6                                & 10         & 251       & 39.46      & 64.71     & 349.0     & 39.955       & 65.912      & 349.0     & 1.2999      \\
7                                & -49        & 224       & 18.41      & 56.87     & 349.0     & 19.591       & 56.603      & 349.0     & 1.2108      \\
8                                & -72        & 73        & 9.97       & 3.96      & 349.0     & 10.619       & 3.778       & 349.0     & 0.6740      \\
9                                & -142       & 303       & -11.91     & 84.29     & 349.0     & -11.613      & 84.501      & 349.0     & 0.3643      \\
10                               & -62        & 154       & 14.58      & 29.23     & 349.0    & 14.633       & 32.121      & 349.0     & 2.8915      \\ 
                                 &            &           &            &           &            &              &             & CR        & 1.4668      \\
                                 &            &           &            &           &            &              &             & TRE       & 1.6887      \\ \hline
\end{tabular}
\end{table}

The metric used to quantify the precision of a calibration method is the calibration reproducibility (CR) \cite{carbajal2013improving}

\begin{equation}
CR=\frac{1}{N} \sum_{n=0}^{N-1}\left|T P_{I, n}-P_{R, n}\right|
\end{equation}
Target registration error (TRE) is also used to evaluate the accuracy of calibration\cite{najafi2014closed}

\begin{equation}
TRE=\sqrt{\frac{1}{N} \sum_{n=0}^{N-1}(TP_{I, n}-P_{R, n})^{2}}
\end{equation}

\begin{figure}[h]
    \centering
    \includegraphics[width=0.7\linewidth]{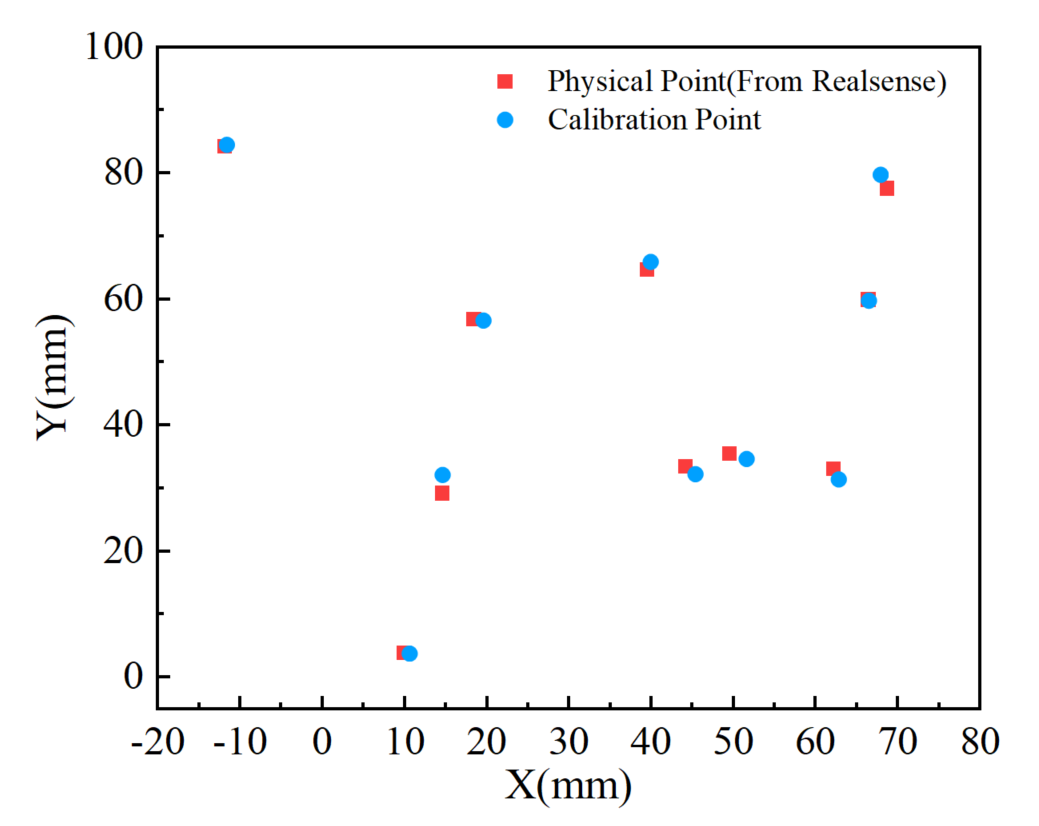}
    \caption{Results between Physical Point and Calibration Point } \label{Results}
\end{figure}

In the experiment, several sets of test points were selected from the collected images and needle coordinates, and the remaining points were used to calculate the transformation matrix $T$. The quantitative analysis of US image calibration is as shown in table \ref{tab1} and Fig.\ref{Results}. Compared with other calibration methods, the comparison results are shown in the table \ref{tab2}.
\begin{table}[h]
\centering
\caption{Comparison with other calibration methods}\label{tab2}
\tabcolsep=0.62em
\renewcommand\arraystretch{1.2}
\begin{tabular}{lcc}
\hline
\multicolumn{1}{c}{Mthod} & \multicolumn{2}{c}{Precision and accuracy of calibration} \\
                          & CR(mm)                      & TRE(mm)                     \\ \hline
N-wire phantom            & 1.97                        & 2.06                        \\
Multi-wedge phantom       & 1.58                        & 1.80                        \\
proposed method           & 1.47                        & 1.69                        \\ \hline
\end{tabular}
\end{table}

\section{Discussion and Conclusion}
In this paper, we propose a fast calibration method for US images based on a depth camera that achieves an accuracy of within 2$mm$. We completed the calibration on the designed experimental platform by solving the intrinsic matrix of the US system in conjunction with the conversion between coordinate systems.

In addition, the feasibility of the proposed method is discussed and verified through experiments. Specifically, in our proposed method, the CR and TRE of the whole system are only 1.4668 mm and 1.6887 mm, respectively. Those errors are smaller than the CR (1.97 mm) and TRE (2.06 mm) of the N-wire phantom calibration method \cite{carbajal2013improving} and the CR (1.58 mm) and TRE (1.80 mm) of the multi-wedge phantom calibration method in \cite{najafi2014closed}. Our proposed calibration method is significantly better than the other two image-based methods in terms of the calibration accuracy of the whole system. 

It should be noted that our method does not require additional calibration molds, but still requires a sufficiently high accuracy of the depth camera. Therefore, similar US calibration methods can be further developed to improve the flexibility and accuracy of US image calibration. For example, a higher precision positioning device could be used to locate the needle tip position, or the overall calibration device could be mounted on a robot to automate and streamline the calibration process in specialized application scenarios. This work provides a reference for the calibration of the US puncture robot as shown in Fig.~\ref{classification}.


%
%
%
%
\bibliographystyle{style}
\bibliography{reference}
\end{document}